\documentclass[journal,11pt]{IEEEtran}
\pdfoutput=1

\usepackage[utf8]{inputenc}
\usepackage{algorithm}
\usepackage{algorithmic}
\usepackage{amsmath}
\usepackage{graphicx}
\usepackage{extarrows} 
\usepackage{comment}
\usepackage[subrefformat=parens,labelformat=parens]{subfig}
\usepackage{bm}
\usepackage{multirow}
\usepackage{threeparttable,booktabs}
\usepackage{tikz}
\usepackage{balance}
\usepackage{courier}                                       
\usepackage[mathcal]{eucal}
\usepackage{filecontents}                                  
\usepackage{pgfplots}
\usepackage{pgfplotstable}
\usepackage{graphicx}                                      
\usepackage{amssymb}
\usepackage{amsfonts}
\usepackage{booktabs}
\usepackage{multirow}
\usepackage{color}
\usepackage[square, comma, sort&compress, numbers]{natbib}
\usepackage{comment}                                       
\usepackage{soul}                                          
\soulregister\cite7
\soulregister\ref7
\soulregister\pageref7
\usepackage{amsthm}
\usepackage{etoolbox}                                      
\usepackage{url}
\usepackage{nth}                                           
\usepackage{bm}                                            
\usepackage{courier}
\usepackage{balance}
\usepackage{threeparttable}
\usepackage[bookmarks=false]{hyperref}                     %
\hypersetup{
    colorlinks = true,
    citecolor  = blue,
    linkcolor  = blue,
    urlcolor   = blue,
}
\usepackage{filecontents}                                  
\usepackage{tikz}
\usetikzlibrary{positioning, fit}
\usepackage{pgfplots}
\usepackage{pgfplotstable}
\pgfplotsset{compat=newest}
\usepackage{caption} 
    
\usepackage{tabu}
\usepackage{tabularx}
\usepackage{rotating}
\usepackage{supertabular}
\usepackage{tikz}
\usepackage{hhline}
\usepackage{longtable}
\usepackage{svg}
\usepackage{rotating}
\usepackage{xcolor}
\usepackage{bbding}

\usepackage{color}
\usepackage{xcolor}
\usepackage{threeparttable}
\usepackage{utfsym}
\usepackage{url}
\usepackage{hyperref}
\usepackage{cleveref} 
\usepackage{multirow}
\usepackage{subfig}
\usepackage{float}
\usepackage{algorithm}
\usepackage[export]{adjustbox}
\usepackage[tableposition = top]{caption} 
\usepackage{booktabs}                     
\usepackage[per-mode = symbol]{siunitx}   
\usepackage{colortbl,booktabs}
\usepackage{multicol}
\usepackage{multirow} 
\usepackage{multirow, makecell, caption}
\usepackage{diagbox}

\usepackage{amsmath,amssymb,amsfonts}
\usepackage{graphicx}
\usepackage{textcomp}
\usepackage{xcolor}

\usepackage{lettrine}

\graphicspath{{./Images/}}

\definecolor{CUHKorange}{RGB}{244,106,18} 
\definecolor{CUHKblue}{RGB}{0,111,190}    
\definecolor{CUHKgreen}{RGB}{0,127,128}   
\definecolor{CUHKred}{RGB}{228,46,36}     
\definecolor{CUHKyellow}{RGB}{198,148,34} 
\definecolor{CUHKdark}{RGB}{114,44,114}   
\definecolor{CUHKmiddle}{RGB}{144,44,144} 
\definecolor{CUHKlight}{RGB}{167,44,167} 

\makeatother

\RequirePackage[normalem]{ulem} 
\RequirePackage{color}\definecolor{RED}{rgb}{1,0,0}\definecolor{BLUE}{rgb}{0,0,1} 

\graphicspath{{./figs/}}

\begin{document}

\title{\textbf{PDNNet: PDN-Aware GNN-CNN Heterogeneous Network for Dynamic IR Drop Prediction }}

\author{
Yuxiang Zhao,
Zhuomin Chai,
Xun Jiang,
Yibo Lin~\IEEEmembership{Member,~IEEE},
Runsheng Wang,
Ru Huang~\IEEEmembership{Fellow,~IEEE}
\thanks{ This work was supported in part by the National Key Research and Development Program of China (No. 2021ZD0114702), 
the Natural Science Foundation of Beijing, China (Grant No. Z230002), the National Science Foundation of China (No. 62034007), 
the 111 Project (B18001).}
\thanks{Y.~Zhao and X.~Jiang are with the School of Integrated Circuits, Peking University.}
\thanks{Z.~Chai is with the School of Physics and Technology and the School of Microelectronics, Wuhan University, the School of Integrated Circuits, Peking University, and the Hubei Luojia Laboratory.}
\thanks{Y.~Lin, R.~Wang and R.~Huang are with the School of Integrated Circuits, Peking University, Beijing, China, 
Institute of Electronic Design Automation, Peking University, Wuxi, China, 
and Beijing Advanced Innovation Center for Integrated Circuits, Beijing, China.}
\thanks{Corresponding author: Yibo Lin (yibolin@pku.edu.cn)}
}

\maketitle
\thispagestyle{empty} 

\begin{abstract}

IR drop on the power delivery network (PDN) is closely related to PDN's configuration and cell current consumption.
As the integrated circuit (IC) design is growing larger, dynamic IR drop simulation becomes computationally unaffordable and machine learning based IR drop prediction has been explored as a promising solution.
Although CNN-based methods have been adapted to IR drop prediction task in several works, the shortcomings of overlooking PDN configuration is non-negligible.
In this paper, we consider not only how to properly represent cell-PDN relation, but also how to model IR drop following its physical nature in the feature aggregation procedure.
Thus, we propose a novel graph structure, PDNGraph, to unify the representations of the PDN structure and the fine-grained cell-PDN relation.
We further propose a dual-branch heterogeneous network, PDNNet, incorporating two parallel GNN-CNN branches to favorably capture the above features during the learning process.
Several key designs are presented to make the dynamic IR drop prediction highly effective and interpretable.
We are the first work to apply graph structure to deep-learning based dynamic IR drop prediction method.
Experiments show that PDNNet outperforms the state-of-the-art CNN-based methods and achieves $545\times$ speedup compared to the commercial tool, which demonstrates the superiority of our method.

\end{abstract}






 


\begin{figure}[t]
\begin{center}
\setlength{\tabcolsep}{1.5pt}
\scalebox{1}{
\begin{tabular}[b]{c c}

\includegraphics[height=.18\textwidth,valign=t]{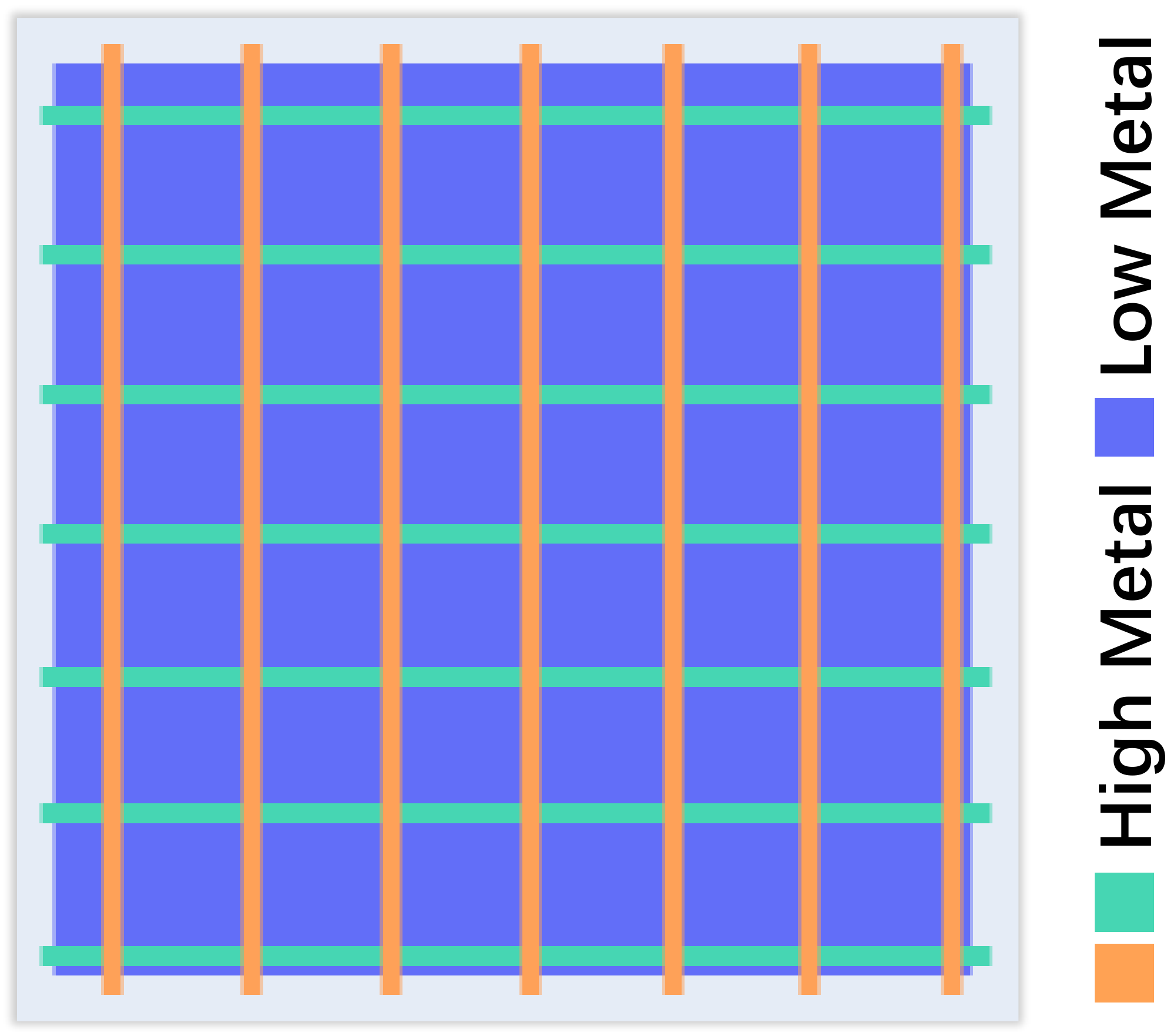}  \ \ &   
\includegraphics[height=.18\textwidth,valign=t]{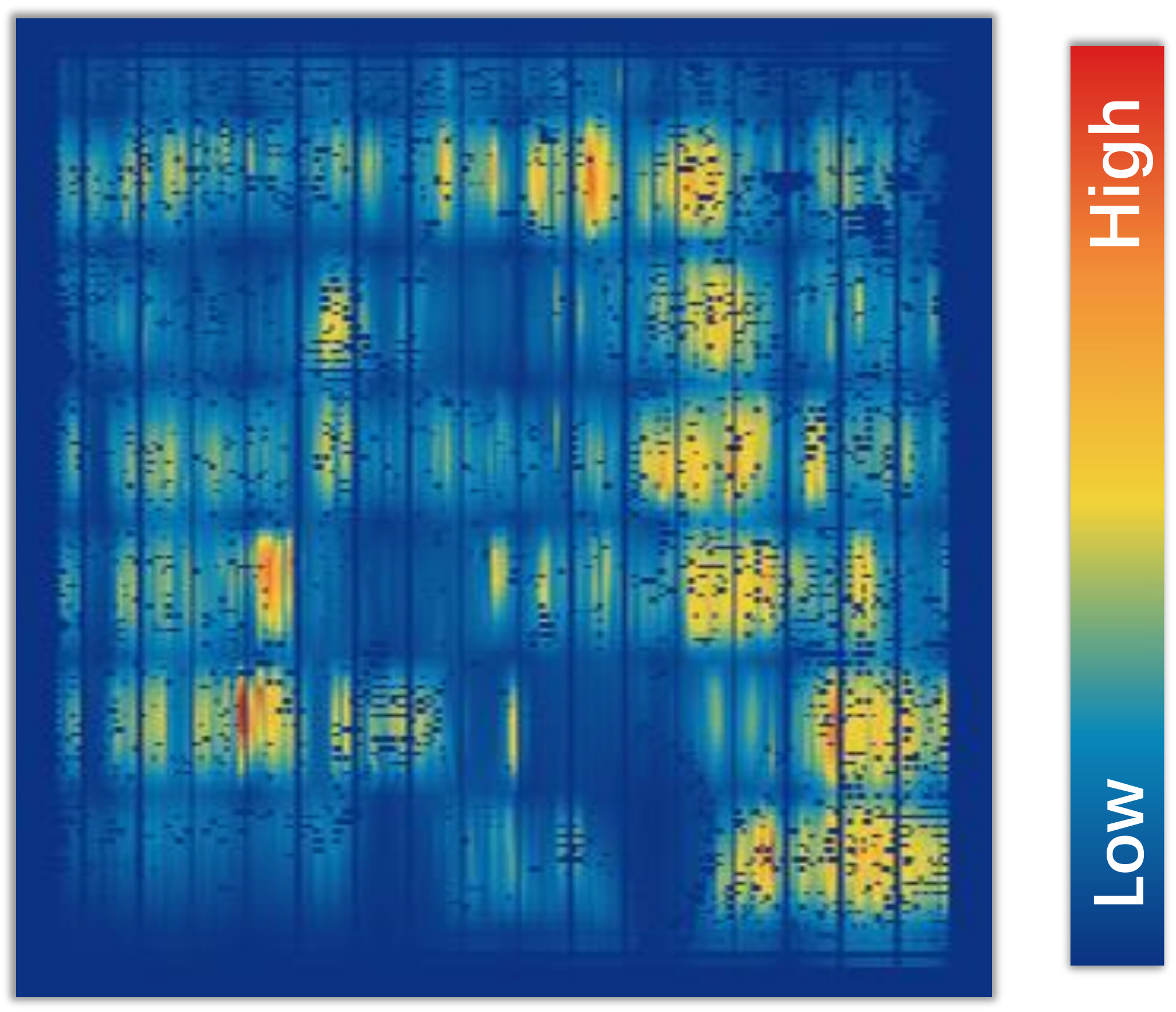}
\\

\vspace{-1mm} 
\\
\multicolumn{2}{c}{\small(a) PDN with sparse grid design.} 
\vspace{3mm} 
\\
\includegraphics[height=.18\textwidth,valign=t]{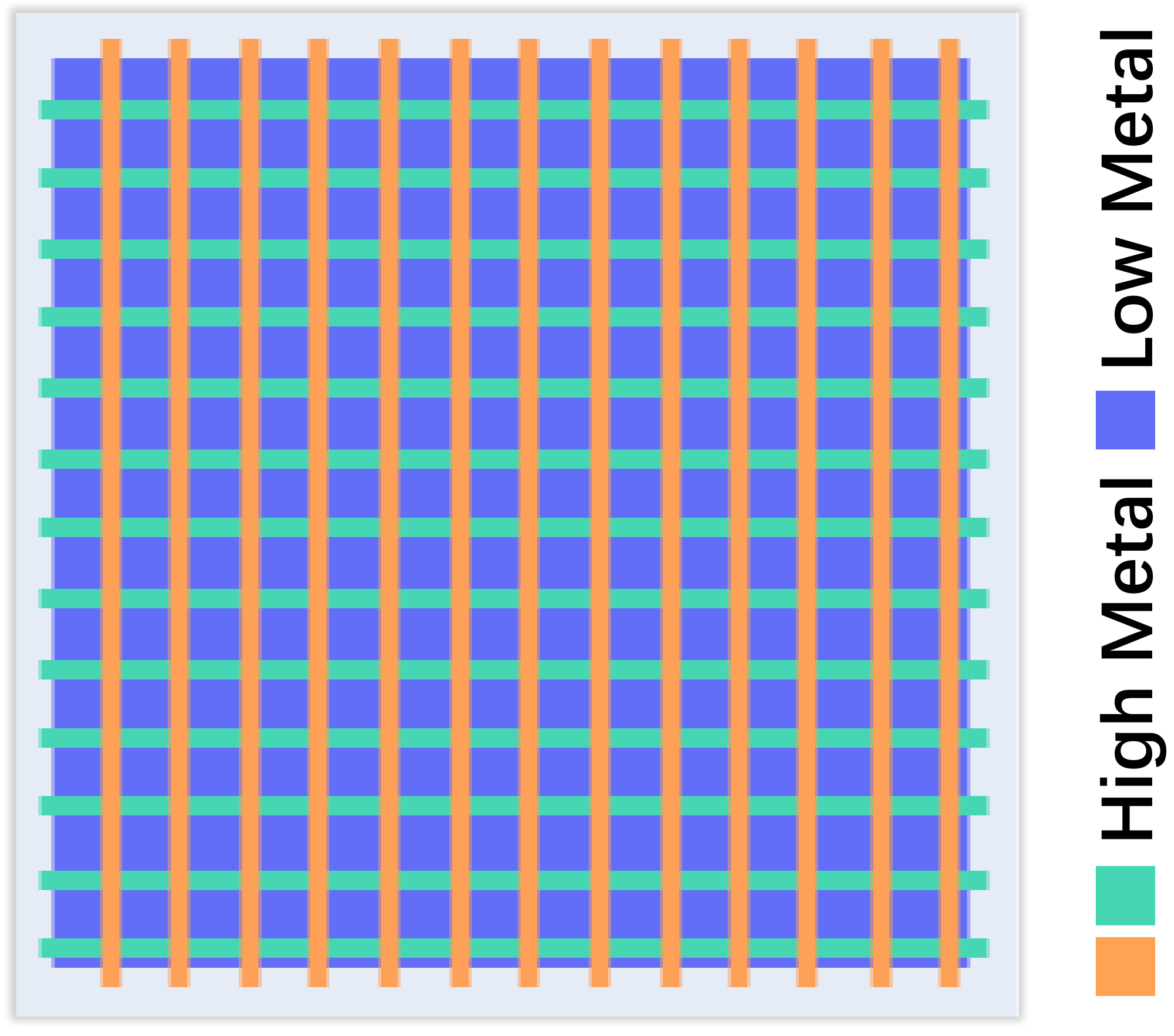}  \  \ &   
\includegraphics[height=.18\textwidth,valign=t]{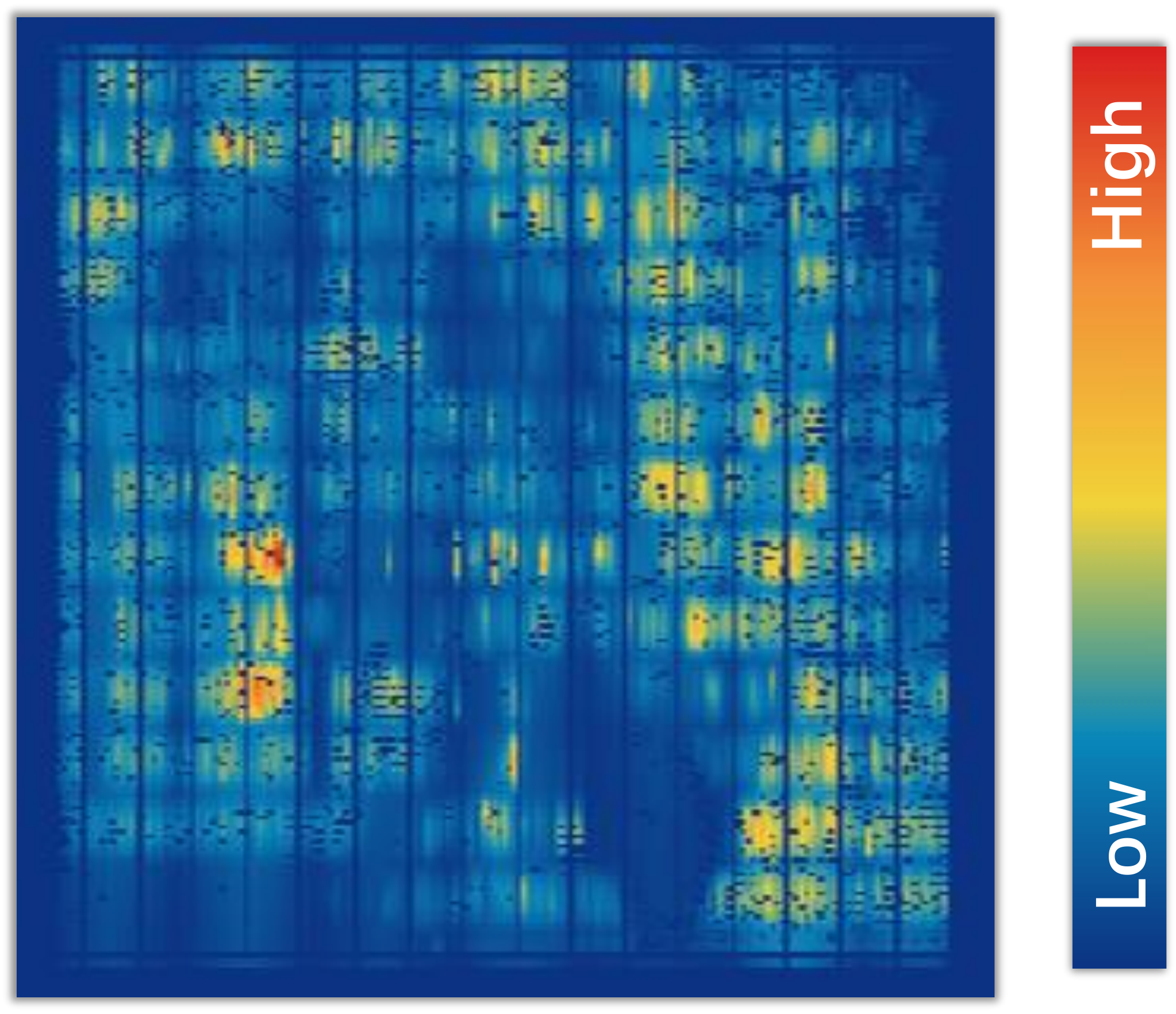}

\\
\multicolumn{2}{c}{\small(b) PDN with dense grid design.}
\vspace{5mm} 
\end{tabular}}
\end{center}
\caption{Example of IR drop maps with different PDNs under the same current load. The left part is PDN architecture. High metal (\emph{e.g.}, \textcolor[RGB]{253, 165, 80}{M8} and \textcolor[RGB]{70, 214, 178}{M7}) and low metal (\emph{e.g.}, \textcolor[RGB]{96, 112, 245}{M1}) are distinguished with different color. The corresponding IR drop map is in the right position. The color approaching \textcolor[RGB]{208, 36, 22}{Red} denotes a higher IR drop value, and approaching \textcolor[RGB]{12, 53, 128}{Blue} denotes lower.
}
\label{fig:instance}
    \end{figure}


\vspace{-0.5em}
\section{Introduction}
\label{sec:intro}
\lettrine[lines=2]{P}{ower} delivery network is responsible for transmitting voltage and current from the chip's IO pins to every transistor ~\citep{power-grid, power-grid-design, pdn-analysis, power-grid-benchmark, power-grid-analysis}.
It contains a hierarchical architecture with a set of power bumps and multi-redistribution layers from top to bottom.
Due to the existence of parasitics, a voltage drop is induced when the supply current flows through PDN.
This common but unpleasant phenomenon is named IR drop, which is determined by two key factors (current load patterns as well as PDN architecture).
The~\Cref{fig:instance} \textcolor{blue}{(a)}\textcolor{blue}{(b)} shows two IR drop distribution maps in distinctive PDN architectures under the same current load pattern.
Large IR drops can hurt chip performance, timing, and in the worst case, its functionality.
Thus, IR drop analysis is a crucial step to verify the satisfaction and reliability of PDN before manufacturing.~\citep{gate-resizing, cell-option, irdrop-dram, irdrop-reduction, co-optimization, irdrop-compensations}
As simulation-based analysis is too expensive, especially for dynamic IR drop, machine learning based IR drop prediction have been achieved to accelerate design closure. 

In literature, convolutional neural networks (CNNs) in machine learning have dominated this task for a long time.
Current CNN-based methods regard dynamic IR drop prediction as an image generation process  \citep{cycle, pix2pix, semantic-image, stargan, condition-gan, fast-irdop-pred}.
%
Mozaffari et al. \citep{mozaffari_efficient_2019} develop an on-chip power supply noise evaluation method, leveraging the CNN model to handle global and local features in prediction.
Xie et al. \citep{xie_powernet_2020} propose a max-CNN structure, called PowerNet, that encodes input power features to capture the peak voltage drop along the temporal axis. 
Chhabria et al. \citep{chhabria_mavirec_2021} propose MAVIREC, which formulates the prediction task as an image-to-image translation problem, utilizing a UNet architecture to percept dynamic current variation.

Although these methods achieve good performance, the prominent drawbacks can not be neglected.
As~\Cref{fig:instance} shows, current load patterns and PDN architectures are two indispensable parts that jointly impact the final IR drop distribution.
Study \cite{static-irdrop} uses an image-like PDN density map to represent cross-grained PDN structure for static IR drop prediction. 
However, in the dynamic IR drop prediction field, conventional works focus on extracting the representation from current load patterns, while almost none of them devote attention to exploring the PDN structures.
The reason mainly lies in two-fold.
Firstly, image-like input features cannot fully represent PDN structure and dynamic current flow, especially powerless when facing the fairly regular PDN mesh (\emph{i.e.} PDN density map has the same value on each pixel in this case). 
Secondly, as pure CNN-based models lack topology relation aggregation approaches, they are neither capable of perceiving fine-grained cell-PDN connection relation, nor modeling the current flowing behavior on PDN.

This paper aims to target IR drop prediction from a brand-new angle based on the physical nature of this problem.
The main contributions of this work are summarized below: 
%

\begin{table*}[tbh]

    \centering
    \caption{Comparison between various dynamic IR drop analysis methods.}
    \vspace{0.3em}
    \label{tab:method-comp}
    \scalebox{1.1}{
    \begin{tabular}{c|c c c c c}
    \toprule[0.12em]
        Name & Method & Model & Cell-PDN Relation & Generality & Speed \\ \midrule
        Voltus~\citep{TOOL_innovus} & Simulation & 
        IR Drop Analysis
        & Yes & High & Medium \\ 
        \midrule
        \citep{Vector-based-DynamicIRdrop2022} & Prediction  & XGBoost & No & Medium & Medium \\ 
        PowerNet~\citep{xie_powernet_2020} & Prediction  & CNN & No & Medium & Medium \\ 
        MAVIREC~\citep{chhabria_mavirec_2021} & Prediction & CNN & No & High & High \\ 
        Our PDNNet & Prediction & GNN-CNN Heterogeneous & Yes (use PDNGraph) & High & High \\ \bottomrule[0.12em]
    \end{tabular}}
\end{table*}
 

\begin{itemize}
    \item We propose a novel graph structure, \textbf{PDNGraph}, which comprehensively represents the fine-grained PDN structure and cell-PDN relation.
    To the best of our knowledge, we are the first to introduce graph structure into deep-learning based dynamic IR drop prediction. 
    \item We propose a novel dual-branch network, \textbf{PDNNet}, that concentrates on two core perspectives.
    The GNN branch tailors the aforementioned PDNGraph from the perspective of PDN awareness.
    The parallel CNN branch focuses on capturing dynamic IR drop variation along the temporal axis.
    \item We conduct extensive experiments on a large-scale dataset. PDNNet outperforms the state-of-the-art CNN-based methods by up to \textbf{24.3\%} reduction in prediction error and achieves \textbf{545$\times$} speedup compared to the commercial tool, which demonstrates the superiority of our proposed method.
    
\end{itemize}

The rest of the paper is organized as follows. 
The IR drop analysis procedure, problem formulation and method comparison are described in~\Cref{sec:preliminary}.
The details of PDNGraph construction and PDNNet architecture are presented in~\Cref{sec:method}.
In~\Cref{sec:experiments}, a series of experiments are conducted to comprehensively explain the effectiveness of PDNGraph and PDNNet.
The paper is concluded with a summary in~\Cref{sec:conclusion}.

\section{Preliminary}\label{sec:preliminary}

\subsection{IR Drop Analysis} \label{sec:preliminary:idrop}
\begin{figure}[h]
    \centering
    \includegraphics[width=0.39\textwidth]{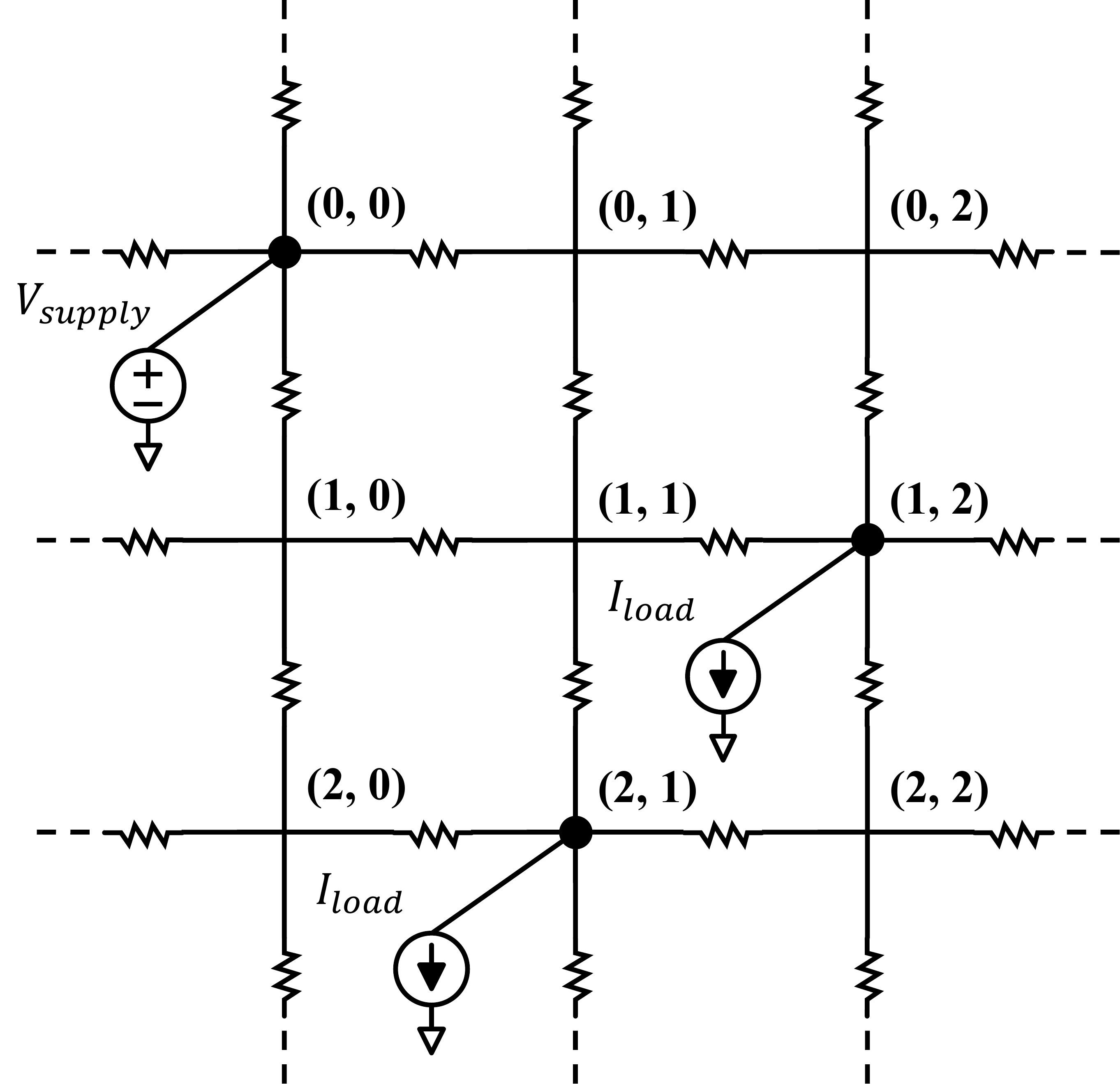}
    \vspace{0.8em}
    \caption{The illustration of PDN grid in IR drop analysis. }
    \label{fig:pdn-grid}
\end{figure}

IR drop analysis simulates  the voltage drop under the combined effect of parasitic and current flow through the PDN \citep{fast-irdrop2}.
Generally, PDN is casted as a grid topology network with parameters, that mainly include impedance and conductance.\citep{10323865, 9358096, 6509647, 9643489, 6657051, 7398003, 6617720, 1487560}
Voltage sources (\emph{i.e.}, power supply rails) and current loads (\emph{i.e.}, cells) respond to simulate the current flowing on the network.
The illustration of PDN grid depicts in~\Cref{fig:pdn-grid}.

\begin{figure}[!ht]
\begin{center}
\vspace{1mm} 
\setlength{\tabcolsep}{1.5pt}
\scalebox{1}{
\begin{tabular}[b]{c}

\includegraphics[width=.45\textwidth,valign=t]{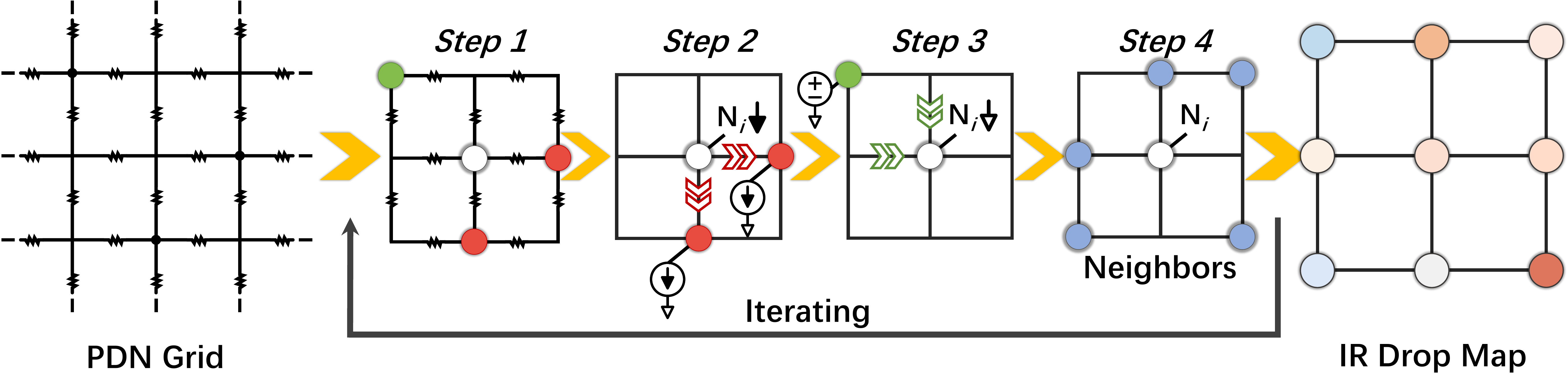} 
\vspace{3mm} 
\\
\small(a) Simulation-based IR drop analysis pipeline.
\vspace{3mm} 

\\
\includegraphics[width=.45\textwidth,valign=t]{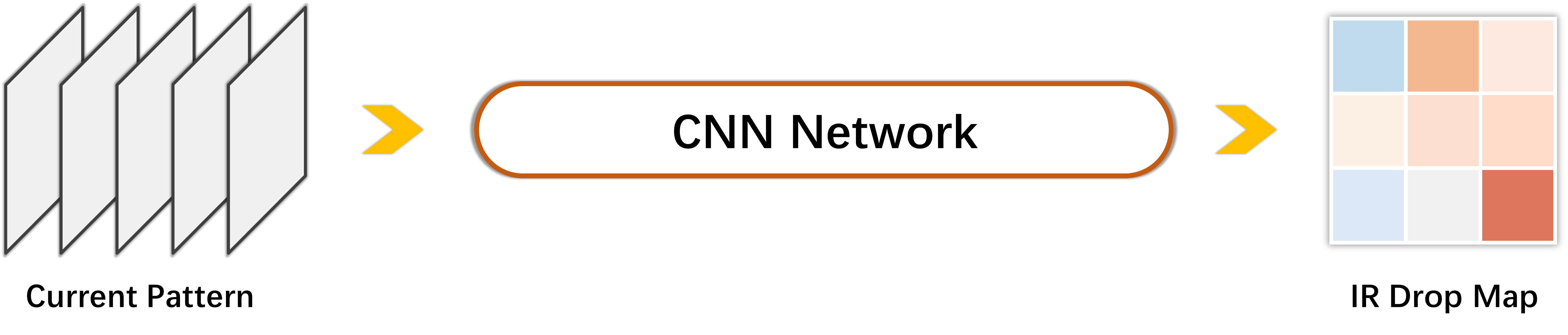}
\vspace{3mm} 
\\
\small(b) Conventional CNN-based IR drop prediction method.
\vspace{3mm} 
\\
\includegraphics[width=.45\textwidth,valign=t]{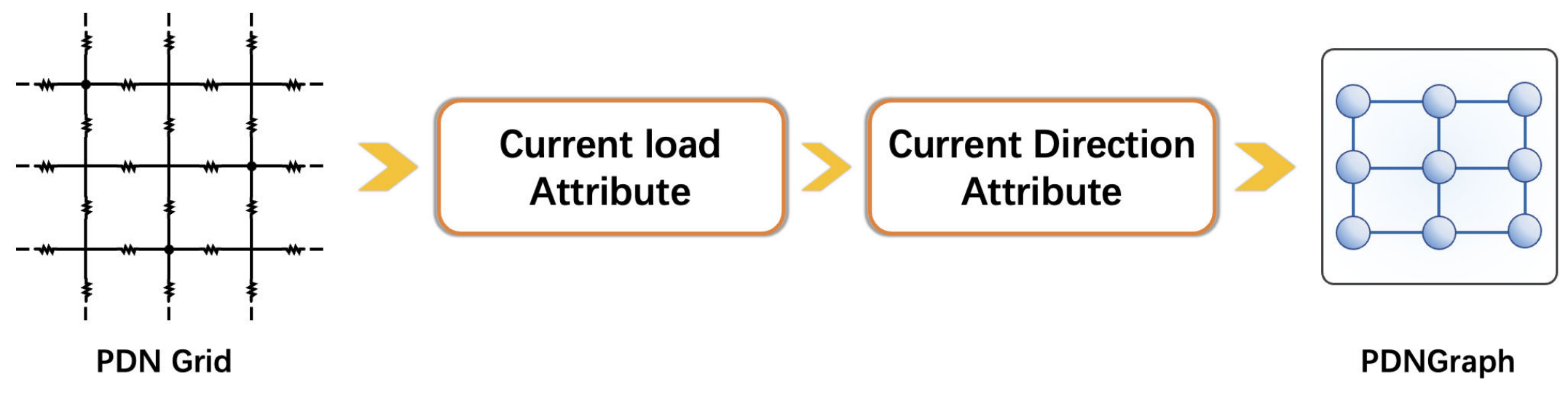}
\vspace{3mm} 
\\
\small(c) Proposed \emph{PDNGraph} construction process.
\vspace{3mm} 
\\
\includegraphics[width=.45\textwidth,valign=t]{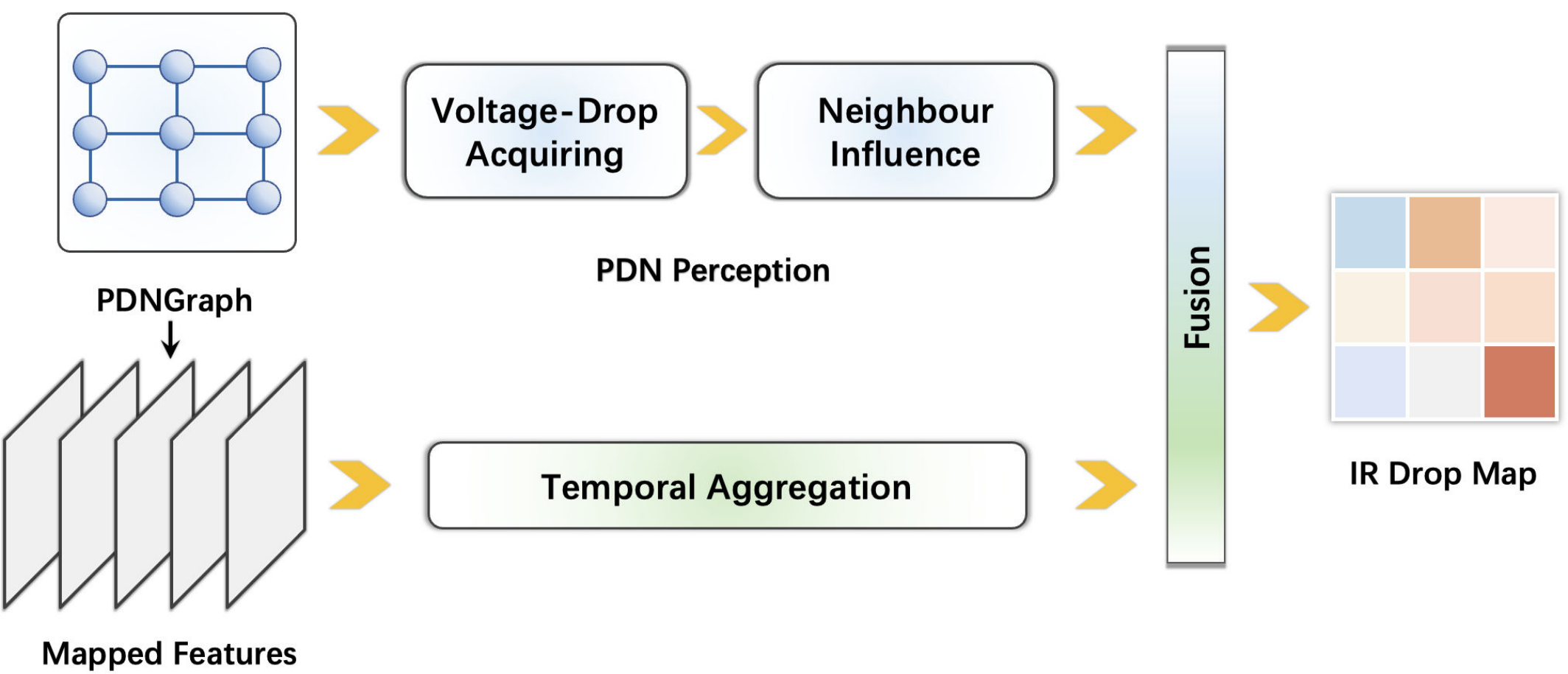}
\vspace{3mm} 
\\
\small(d) Proposed \emph{PDNNet} IR drop prediction framework.
\vspace{3mm} 
\end{tabular}}
\end{center}
\caption{Analysis of simulation-based method, conventional deep-learning prediction methods, our proposed PDNGraph, and PDNNet framework.
}
\label{fig:pipeline}
\end{figure}

Hence, the IR drop analysis involves the solution of a large system \citep{fast-irdrop} of equations of the form,

\begin{equation}
GV \ =\  J
\end{equation}

\noindent where $G$ is the conductance matrix for the PDN, $J$ is the vector of current sources, and $V$ is the set of voltages at each node in the PDN.
The whole procedure~\citep{fast-irdrop} iteratively measures voltage at each node in four key consecutive steps as depicted in~\Cref{fig:pipeline} \textcolor{blue}{(a)}. 
\begin{itemize}
    \item \emph{Step 1}: Calculate effective resistance and capacitance between this node, the on-grid voltage sources, and current loads parts.
    \item \emph{Step 2}: Calculate node's voltage drop induced by global current flows to current loads.
    \item \emph{Step 3}: Acquiring node's voltage supply transmitted by current flows from voltage source.
    \item \emph{Step 4}: Consider node's voltage fluctuation influenced by the voltage of neighbor nodes.
    
\end{itemize}
The IR drop analysis can thus be formulated as:

\begin{equation}
y\ = f \circ g \circ h \circ s(x)
\end{equation}

where $s$ present voltage, current map and effective resistance calculation, $h$ denotes the voltage drop acquiring, $g$ represents the calculation of voltage source supply, and $f$ denotes the influence from neighbor.
Notation $\circ$ represents the connections between two operations.
The analysis formulation explicitly indicates that PDN structure and dynamic current pattern variations jointly determine IR drop distribution map on the grid.


\begin{figure*}[t]
\label{table:mehod-comparsion}
    \centering
    \includegraphics[width= 0.98\textwidth]{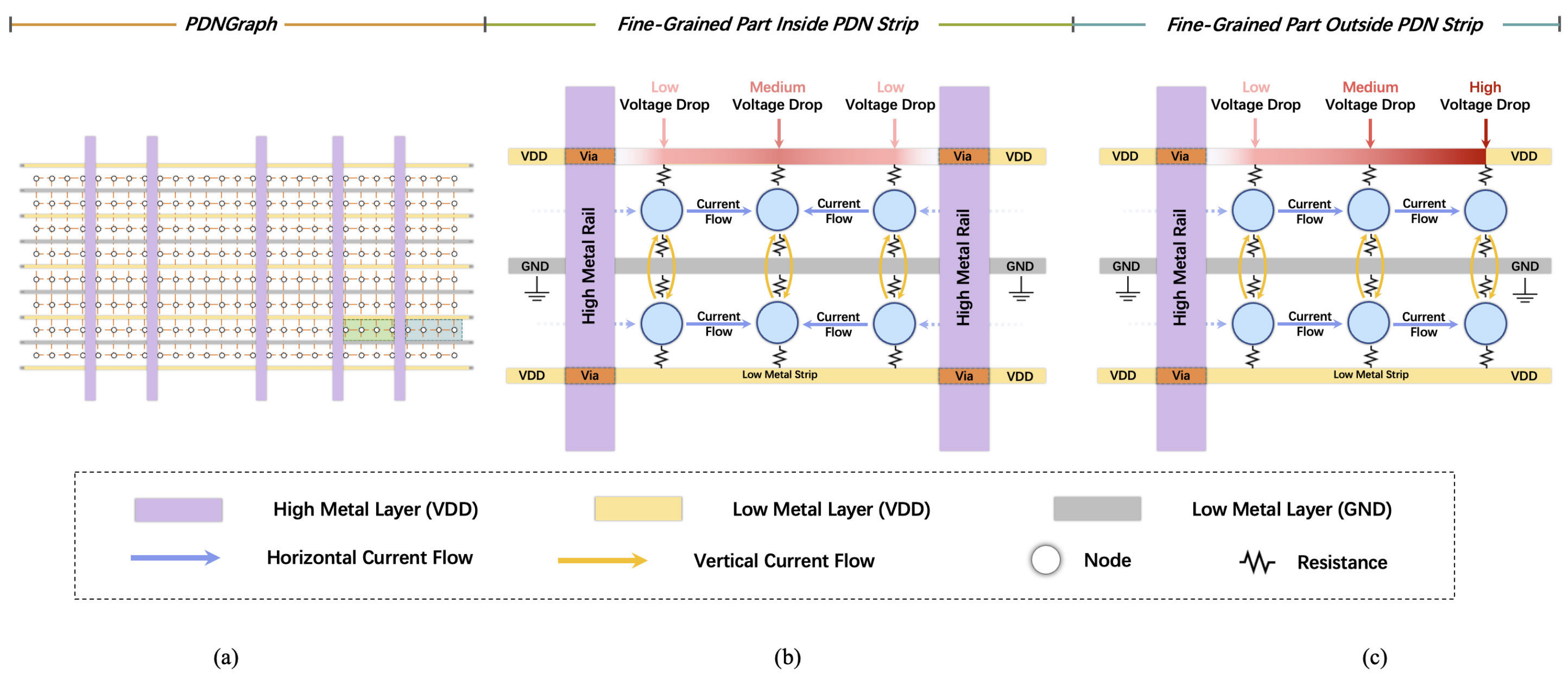}
    \vspace{0.3em}
    \caption{Illustration of our PDNGraph construction. Figure better viewed in color mode. From left to right is (a) The whole graph on IC layout. (b) The fine-grained part inside PDN strip. (c) The fine-grained part outside PDN strip.}
    \label{fig:pdngraph}
\end{figure*}


\subsection{Problem Formulation}
The former IR drop modeling suffers from a high computation burden.
Hence, we approach this procedure in a learnable fashion.
This task, named IR drop prediction, is of great value in accelerating loops of IC design closure.
We identify our learning formulation as follows: 

\textbf{Problem 1.} \emph{Given a series of data after placement $\{PL\}$, consisting of PDN attributes $P=\{p_i \mid 1\leq i \leq N_p \}$, standard cells locations $C=\{c_i \mid 1\leq i \leq N_c \}$ and cell current related features $X=\{x_i^k \mid 1\leq i \leq N_c, 1\leq k \leq K \}$, our object is to design an algorithm $\mathcal{G}$ to compose these features in a specific manner and propose a learning function $\mathcal{F}$ to intake the processed data to generate an accurate dynamic IR drop distribution map.}

\subsection{Comparison with Other Methods.}
\Cref{tab:method-comp} presents the difference between PDNNet, Voltus, and other typical models for dynamic IR drop analysis.
Voltus is a commercial tool that employs simulation-based method to derive an IR drop map.
The results obtained from this tool serve as our ground reference.
Within prediction-based methods, XGBoost and CNN represent two commonly employed models in machine learning and deep learning, respectively.
Cell-PDN relation refers to the consideration of the physical interconnection between the PDN network and cells.
Generality and speed stand as two pivotal parts of the practical usability of these methods. The results are based on our experiments in~\Cref{sec:experiments}.
The PDNGraph data structure and PDNNet model enable our method to generate an accurate IR drop map by mimicking the dynamic current flow that exists in PDN structure, distinguishing it from the abovementioned prediction-based methods.
Specifically, PDNGraph uses graph structure to favorably represent the fine-grained PDN structure and intrinsic cell-PDN physical relation.
In the meantime, PDNNet leverages a dual-branch GNN-CNN heterogeneous network to achieve a comprehensive perception of both PDN architecture and dynamic IR drop variation.
These advancements boost our PDNNet's performance significantly.

\section{Method}\label{sec:method}
In this section, we propose a solution pipeline, including a novel graph structure, \emph{PDNGraph}, and a novel network architecture, \emph{PDNNet}.
The overview of PDNGraph and PDNNet is shown in~\Cref{fig:pipeline} \textcolor{blue}{(c)} and~\Cref{fig:pipeline} \textcolor{blue}{(d)}. 

\begin{figure*}[!t]
    \centering
    \includegraphics[width= 0.88\textwidth]{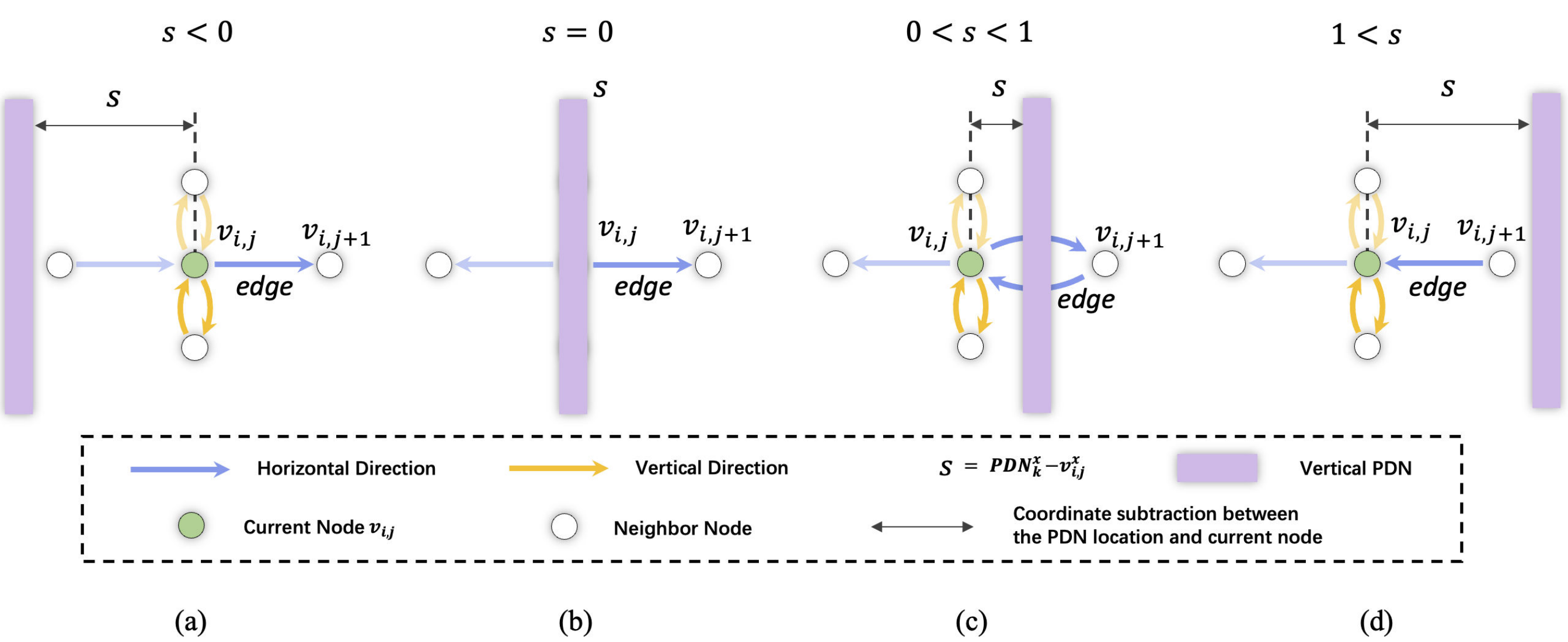}
    \vspace{0.3em}
    \caption{The figure shows how the directions of edges between nodes in PDNGraph are determined. We categorize the relationship between the PDN location and the current node into four cases.}
    \label{fig:pdn_cell}
\end{figure*}

For PDNGraph, it is a directed graph that encompasses the electrical properties of PDN structure and current flow.
We assign several current-related features (\emph{e.g.}, leakage power, internal power, switching power) to each node as internal attributes.
Then, the inter-node edge is constructed to direct from the high power rail toward the low redistribution metal strip, which follows the actual current flow direction through PDN.
As such, PDNGraph is built as a bridge to establish the fine-grained cell-PDN connection.

Furthermore, our proposed PDNNet is a dual-branch heterogeneous aggregation network.
Specifically, for the GNN-branch, we employ a graph neural network to aggregate PDNGraph.
We adopt different GNN layers in the voltage-drop acquiring and neighborhood influence stages, respectively, which align more closely with the formulation pipeline in~\Cref{sec:preliminary:idrop}.
As for the CNN-branch, we introduce a 3D encoder and 2D decoder network with several shortcut connections to capture the dynamic IR drop fluctuation.
Finally, a small yet effective fusion model merges each branch output to yield the final prediction result.
We elaborate on the details of several key designs of our method in the following sub-sections.

\subsection{PDNGraph}\label{sec:method:PDNGraph}

To make appropriate input data for accurate dynamic IR drop prediction, there are two main challenges.
Firstly, the current load pattern of cells needs to extract in both the spatial and temporal axis, which is different from other tasks.
Secondly, the PDN network characteristics are also critical to impact the voltage drop distribution, which is a noteworthy clue.
However, the existing methods stress more on simply leveraging current patterns to approach IR drop prediction, while overlooking the intrinsic cell-PDN physical relation and fine-grained structures of PDN.
To counter these limitations, we shed light on characterizing these factors in graph format.
We propose a novel graph data structure, named PDNGraph, to comprehensively represent PDN structure and cell-PDN relationship.
We elaborate PDNGraph attributes as follows.


\begin{algorithm}[h]
\label{algo:pdngraph}
\caption{PDNGraph Construction Algorithm}
\begin{algorithmic}[1]
\REQUIRE Layout dimensions $W$ and $H$, grid $g$ with fixed size ($d^x$, $d^y$), a set of vertical PDN with $x$ axis position $\{PDN^{x}\}$.
\ENSURE Our PDNGraph.

\STATE Divide the layout using grid $g$, $N_w= \frac{W}{d^x}, N_h= \frac{H}{d^y}$.

\FOR {$i = 1$ to $N_h$}
    \FOR {$j = 1$ to $N_w$}
        \STATE Set the initial value $s = \infty$.
        
        \FOR {each $PDN^{x}_{k}$ in $\{PDN^{x}\}$}
            \STATE Compute the coordinate subtraction $s$ between $PDN_{k}$ and current node $v_{i,j}$, $s\_tmp = PDN^{x}_{k} - v_{i,j}^{x}$.
            \IF {$s\_tmp < s$}
                \STATE $s = s\_tmp$
            \ENDIF
        \ENDFOR
        
        \IF {$j < N_w$}
            \IF {$s \leq 0$}
                \STATE Set the edge direction from node $v_{i,j}$ to $v_{i,j+1}$ (the closest $PDN$ is on the left of the current node $v_{i,j}$), edge is $ v_{i,j} \longrightarrow v_{i,j+1}$.
            \ELSIF {$0< s \leq 1$} 
                \STATE Set the bidirectional edge between node $v_{i,j}$ and $v_{i,j+1}$ (the closest $PDN$ is between node $v_{i,j}$ and $v_{i,j+1}$), edge is $ v_{i,j} \longleftrightarrow v_{i,j+1}$. 
            \ELSIF {$1<s$}
                \STATE Set the edge direction from node  $v_{i,j+1}$ to $v_{i,j}$ (the closest $PDN$ is on the right of the current node $v_{i,j}$ and neighbor node $v_{i,j+1}$), edge is $ v_{i,j+1} \longrightarrow v_{i,j}$.
            \ENDIF
        \ENDIF
        
        \IF {$i < N_h$}
            \STATE Set the bidirectional edge between node $v_{i,j}$ and $v_{i+1,j}$, edge is $v_{i,j} \longleftrightarrow v_{i+1,j}$.
        \ENDIF
    \ENDFOR
\ENDFOR
\end{algorithmic}
\end{algorithm}


\subsubsection{\textbf{Graph Construction}}

The~\Cref{fig:pdngraph} \textcolor{blue}{(a)} illustrates our PDNGraph construction outline. 
We regard the circuit layout as a checkerboard, defining
each grid \( g \) that covers multiple cells as a “node” $v$ of our PDNGraph. 
The edges connect adjacent nodes $N_v$ in one-hop and are restricted only in the horizontal and vertical directions, depicted in \Cref{fig:pdngraph} \textcolor{blue}{(b)(c)}.
The edge distances between
nodes implicitly encapsulate the resistance and capacitance characteristics of the power supply metal. 
To ensure uniformity in these attributes, we employ a ``uniform grid" strategy same as \citep{xie_powernet_2020} for constructing the
PDNGraph, which guarantees consistent edge properties across the graph.
Assuming the layout size is $W \times H$, we utilize fixed grids size \(d_x \times d_y \) to segment the layout to $N_w \times N_h $, where $N_w = \frac{W}{d_x}$ and $N_h = \frac{H}{d_y}$.
In PDNGraph, we opt not to explicitly calculate resistance and capacitance for each edge, for fully believing electrical law can be  \emph{``learned''} by an artificial neural network (\emph{e.g.}, our PDNNet).
Hereafter, PDNGraph manifests a consistent manner with the grid network in IR drop analysis.
Note that the grid division is not dependent on the PDN structure and would only affect the edge direction in the PDNGraph that is elaborated in the current direction attribute part.

\begin{figure}[h]
    \centering
    \includegraphics[width= 0.48\textwidth]{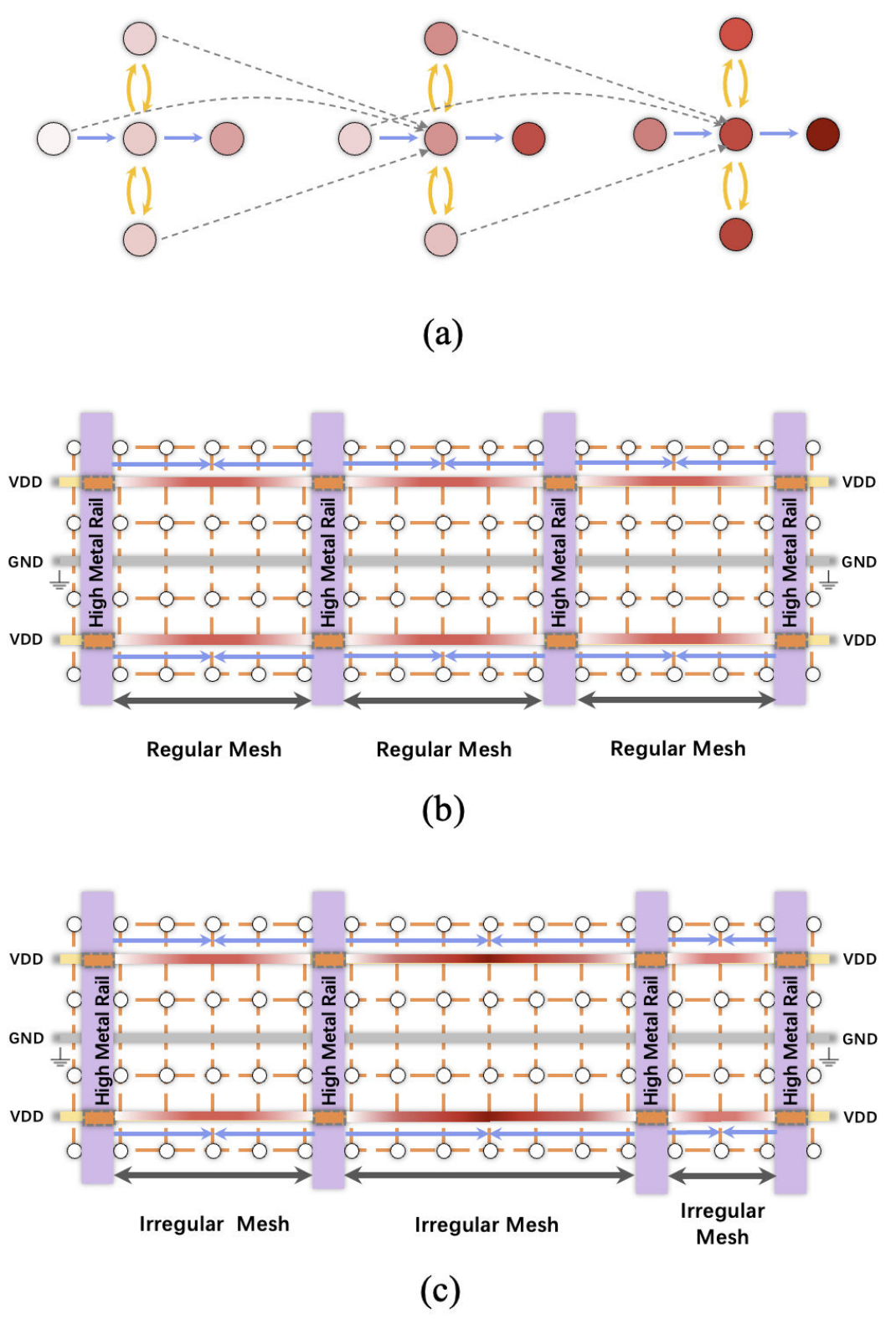}
    \vspace{0.3em}
    \caption{
    (a) message aggregation and passing on PDNGraph. (b) PDNGraph construction of regular PDN grid. (c) PDNGraph construction of irregular PDN grid. Generally, standard cells that are further away from the PDN experience higher IR drop.}
    \label{fig:irr-re}
\end{figure}


\subsubsection{\textbf{Current Load Attribute}}
According to the analysis of IR drop modeling, the cells' current consumption pattern is the key feature of node attributes.
Following the protocol of typical work in \citep{xie_powernet_2020}, as power consumption is proportional to current, PDNGraph also opts for cell power as its input feature.
The encompassed power features are:

\begin{figure*}[t]
    \centering
    \includegraphics[width= 0.95\textwidth]{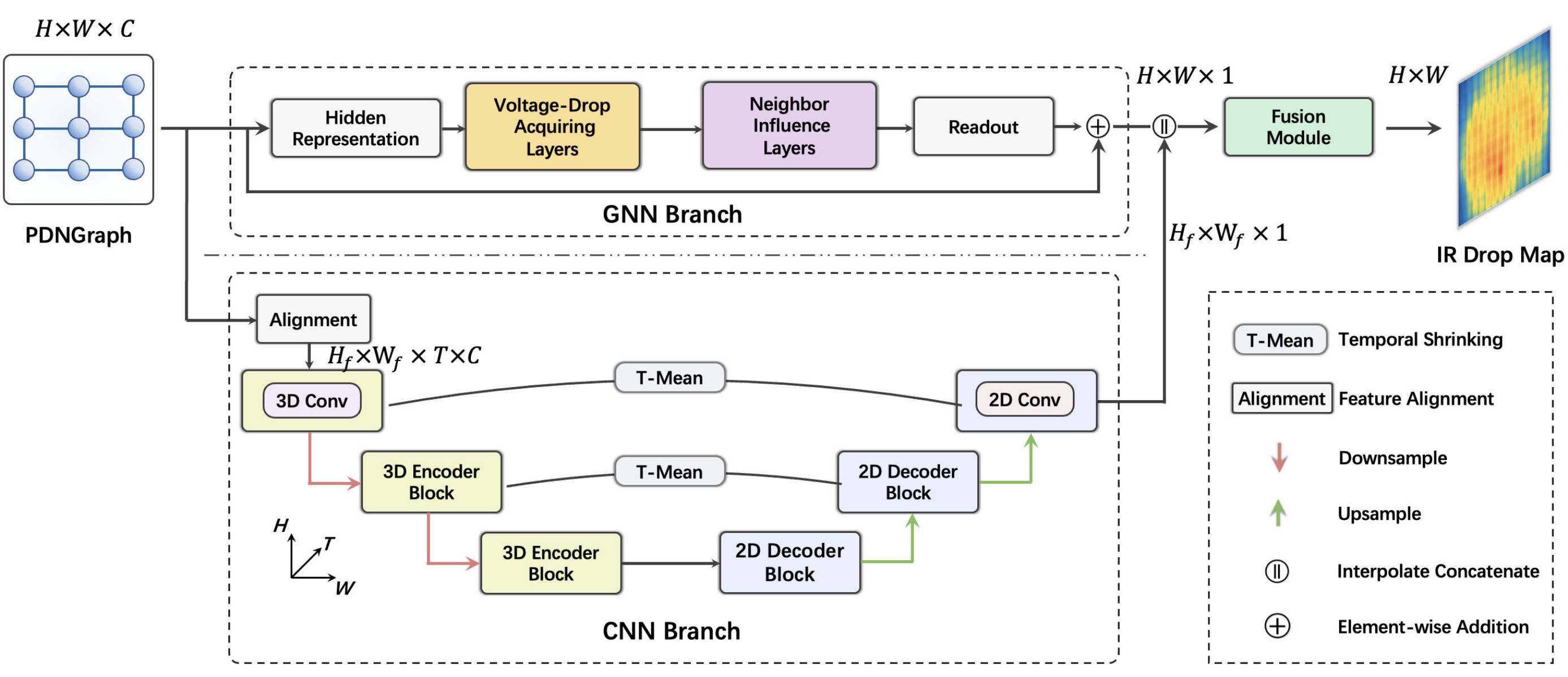}
    \vspace{0.3em}
    \caption{Architecture of PDNNet for dynamic IR drop prediction.
    Our PDNNet consists of two parallel branches: the GNN branch in the upper part and the CNN branch in the lower part. $H, W, T$ denotes the input features’ height, width, and temporal dimension.}
    \label{fig:arch}
\end{figure*}


\begin{itemize}
     \item \textbf{Leakage power,} also known as static power, is the power dissipated when a small current flowing between its source and drain terminals while in the off state.
    \item \textbf{Internal power,} refers to the power consumed by the active components during its operation.
    Since it is variant along temporal, internal power is a type of dynamic power.
    \item \textbf{Switching power,} another type of dynamic power, is the power consumed during the transition of electronic signals switch from one logic level to another (\emph{e.g.}, from 0 to 1).
\end{itemize}

We sum each cell power feature within a gird to form the overall node features $x$, $x\in \mathbb{R}{^{N\times C}}$, where $N$ denotes the node number in PDNGraph, $N = N_w \times N_h $ and $C$ represents the input feature dimension.

\subsubsection{\textbf{Current Direction Attribute}}

In the general PDN configuration, the high power supply VDD rails (HPR) transmit current to the low redistribution VDD layer (LRL).
Despite taking complete PDN into account, we opt for the most relevant part of IR drop distribution to avoid redundancy.
Firstly, without considering the macros, LRL (\emph{i.e.}, M0 or M1) is the layer that directly connects to the cells where most of the IR drop occurs.
Secondly, the location where crisscrossed HPR directly connects to LRL through vias has the least IR drop violation.
To this end, the power supply process can be illustrated that the current flow spread from HPR ends to LRL, along the LRL strip direction towards each node.
The voltage drop at this point, as visually explained in~\Cref{fig:pdngraph} \textcolor{blue}{(b)}\textcolor{blue}{(c)}, also increases gradually when LRL moves away from the HPR.
Therefore, we designated the edges' direction oriented away from the
nearest high-layer power strip to represent the directional flow of current through the PDN structure.

Specifically, as illustrated in lines 4-10 of Algorithm 1, PDNGraph primarily utilizes the smallest coordinate subtraction $s$ between set of vertical PDN's position $\{PDN^{x}\}$ in $x$ axis, and current node position $v_{i,j}^{x}$.
We categorize the construction of graph edges into three cases based on $s$, as illustrate in \Cref{fig:pdn_cell}. \Cref{fig:pdn_cell} (a)(b) corresponds to lines 12-13, \Cref{fig:pdn_cell} (c) corresponds to lines 14-15, \Cref{fig:pdn_cell} (d) corresponds to lines 16-17, and finally, vertical edge is determined as bidirectional corresponds to lines 20-22 in the Algorithm.

That is, the nodes with the same current pattern, may demonstrate distinct IR drop values at different positions among the two high metal rails. 
Since there is no explicit current direction between vertical nodes, we consider them as bidirectional edges.
Further, we neglect the influence of current on the voltage drop across HPR with respect to their low resistance nature.
Note that the edge directions just approximate the current directions. It does not matter that much when the real current directions differ from the edge directions we set, since we allow current values to be negative according to what is learnt from the training data.

Additionally, our PDNGraph can be adaptive to regular and irregular PDN with high flexibility, we illustrate the graph
construction for different types of PDN structures
through figures, shown in~\Cref{fig:irr-re} \textcolor{blue}{(a)(b)}.
Typically, the IR drop value is significantly influenced by the
footprint to the closest high-layer PDN strip. Therefore, areas within a wider PDN pitch are likely to exhibit
higher IR drop values compared to those within a narrower pitch, given identical current consumption.
Consequently, our PDNGraph employs variable edge directions between nodes,
which are determined by the location of HPR to reflect the current flow direction through PDN.

\subsection{PDNNet}
\subsubsection{\textbf{GNN Branch}}

Graph Neural Networks (GNNs) have achieved significant success in node analysis tasks (\emph{e.g.}, node classification, node clustering, etc).
Instead of stacking the same GNN blocks, here we opt for distinct graph aggregation layers following the formulation pipeline in~\Cref{sec:preliminary:idrop}.
The architecture is illustrated in the top of PDNNet in~\Cref{fig:arch}.
In the GNN branch, we first feed the PDNGraph into hidden representations through Multi-Layer Perceptrons ($MLP$).
Then, deeper representation is generated by message-passing through voltage-drop acquiring and neighbor influence layers respectively.
Finally, the GNN branch representation is exported through the readout layers.

\noindent\textbf{Voltage-Drop Acquiring Layer. }
To utilize the current-related features to learn the voltage drop, we stack $l$ voltage-drop acquiring blocks to form this layer.
To better acquire one node $v_{i,j}$'s voltage drop representation, we sum all the neighbors' voltage drop features,
\begin{equation}
m_{v} \ ={\displaystyle \sum _{k\in N_v}\left( v_{k}^{(l-1)}\right)},
\end{equation}
%
\noindent 
Node \( k\) is within the neighborhood \( N_v \) of node \( v\), \( v_k^{(l-1)} \) represents the feature of node \( k\) at layer \( l-1\) and $m_{v}$ is the neighbors' feature summation. Therefore, we update the feature of this node by concatenating all neighbor effects,
%
\begin{equation}
v_{i,j}^{(l)} =f_{\phi }( v_{i,j}^{(l-1)} \ \| \ m_{v}),
\end{equation}
%
\noindent where $\phi$ is the learnable parameters of voltage-drop acquiring blocks and $\ \| \ $ denotes the concatenation function.

\noindent\textbf{Neighbor Influence Layer. }
Since the node features have already captured the characteristics of voltage drop through the above layers, the objective of this layer is to learn and refine the influence of neighbors on each node.
We therefore update the PDNGraph to an bidirected graph while still preserving its topological structure.
Subsequently, learnable function blocks $f_{\theta }$ are employed to update the node representation using both the original node features and the ones from its neighbors,
\begin{equation}
v_{i,j}^{(l)} =f_{\theta }( v_{i,j}^{(l-1)} \ \| \ v_{k}^{(l-1)}) \ \  k\in N_v,
\end{equation}
%
\noindent where $\theta$ is a set of learnable parameters of graph convolution layers.

\noindent\textbf{Voltage Representation Readout. }
After the preceding two layers of message-passing, we employ an $MLP$ layer followed by a $tanh$ activation to yield the PDN-related representation.

\begin{equation}
y^{gnn} =tanh( MLP( x^{(l)} ) ),
\end{equation}

\begin{table}[tb]
\begin{center}
\caption{\small The statistics of CircuitNet. Note that training \& testing datasets have no intersection in designs.}
\label{table:statistics}
\vspace{0.3em}
\setlength{\tabcolsep}{1.5pt}
\scalebox{1.06}{
    \centering
\begin{tabu}{l|c|c|ccc}
\toprule[0.12em]
\multirow{2}{*}{}      & \multirow{3}{*}{Design} & \multirow{3}{*}{\makecell{Number \\ of \\  Samples}} & \multicolumn{3}{c}{Netlist Statistics}                                                                             \\ \cline{4-6} 
                       &                &          & \multicolumn{1}{c}{\#Cells} & \multicolumn{1}{c}{\#Nets} & \begin{tabular}[c]{@{}c@{}}Cell Area\\ ($\mu m^2$)\end{tabular}\\
                       \midrule[0.12em]
\multirow{4}{*}{Train} & \texttt{RISCY-a} & 2003 & \multicolumn{1}{c}{44,836}       & \multicolumn{1}{c}{80,287}      &   65,739  \\
                       & \texttt{RISCY-b} & 1858 & \multicolumn{1}{c}{30,207}   & \multicolumn{1}{c}{58,452}  & 69,779  \\ 
                       & \texttt{RISCY-FPU-a} & 1969 & \multicolumn{1}{c}{61,677}       & \multicolumn{1}{c}{106,429}      &  75,985 \\ 
                       & \texttt{RISCY-FPU-b} &  1248 & \multicolumn{1}{c}{47,130}   & \multicolumn{1}{c}{84,676}  & 80,030  \\ 
                       \midrule
\multirow{2}{*}{Test}  & \texttt{zero-riscy-a} & 2042   & \multicolumn{1}{c}{35,017}   & \multicolumn{1}{c}{67,472}  & 58,631   \\ 
                       & \texttt{zero-riscy-b} & 1122  & \multicolumn{1}{c}{20,350}   & \multicolumn{1}{c}{45,599}  & 62,648  \\ 
    \bottomrule[0.12em]
    \end{tabu}}
\end{center}
\end{table} 

\subsubsection{\textbf{CNN Branch}}

CNN branch in PDNNet is performed as a parallel temporal branch of GNN's.
Although GNN-branch can obtain considerable performance, CNN branch is indispensable for perceiving current alteration along the time axis in dynamic IR drop prediction.
Details can be founded in the experiment~\Cref{sec:experiments}.
To achieve temporal perception, we directly adopt a classical encoder-decoder structure with 2 shortcut connections.
The architecture is present in the bottom of~\Cref{fig:arch}.
More advanced architectures can be applied here, but this is not the focus of this work.
Details of the CNN branch are described as follows.
The intake features share the same as those of the GNN branch but align to a fixed size $H_f \times W_f$.
We map the graph node features onto a 2D space and insert a dummy dimension instead of the original channel.
The transformed feature shape is $x\in \mathbb{R}{^{H_f \times W_f\times T\times C}}$, where $T$ denotes the number of input features and $C$ means feature channels through the representation process.
The input feature is first fed into a 3D encoder network that goes through $l$ encoder layers, each layer comprise of two 3D convolutional and down-sampling blocks.
After feature encoding, the output embedding shape converts to $x^{cnn}\in \mathbb{R}{^{\frac{H_f}{2l}\times \frac{W_f}{2l}\times \frac{T}{l}\times C}}$, where $l$ opt as 3 in our model.
At this point, we shrink the temporal channels since they have been sufficiently extracted.
%
\begin{equation}
x^{cnn}\in \mathbb{R}{^{\frac{H_f}{2l}\times \frac{W_f}{2l} \times \frac{T}{l} \times C }}\ \longrightarrow\  x^{cnn}\in \mathbb{R}{^{\frac{H_f}{2l}\times \frac{W_f}{2l}\times C}}
\end{equation}
%
Subsequently, comprising $l$ layers of deconvolution and up-sampling blocks, the 2D decoder network maintains mirror symmetry in spatial shape with the 3D encoder counterpart at each level.
To further assist the recovery of low-level information from the original feature, we fuse the 4D features, as shown in~\Cref{fig:arch},  from the left side into each layer’s 2D block.
Finally, the CNN network outputs IR drop temporal representation $y^{cnn} \in \mathbb{R}{^{H_f\times W_f \times 1}}$.


\subsection{Fusion Module}
In the last part, we fuse upstream embedding together to get a comprehensive representation.
We first interpolate the feature dimensions from $y^{cnn}\in \mathbb{R}{^{H_f\times W_f\times 1}}$ to $y^{cnn}\in \mathbb{R}{^{H\times W\times 1}}$. Then we flattening the temporal representation $y^{cnn}\in \mathbb{R}{^{H_f\times W_f\times 1}}$ into $y^{cnn}\in \mathbb{R}{^{H_f W_f}}$. 
A concatenation function is imposed to merge these features and pass them through an $MLP$ layer as a regressor and reshape to 2D to generate the final IR drop prediction map.
%
\begin{subequations}
 \begin{align}
z^{prediction}_{1d} \ =\ MLP(  y^{gnn} \ \ \| \  \ y^{cnn}) , \\
z^{prediction}_{2d} \ =\ Reshape(z^{prediction}_{1d}),
\end{align}
\end{subequations}

\subsection{Discussion}\label{sec:method:discussion}
A previous work \citep{static-irdrop} on static IR drop prediction also highlighted the importance of PDN structures in achieving IR drop predictions. The authors transformed the PDN structure into a PDN density map (each pixel denotes the total number of PDN strips covering this location) and an effective distance map from the power supply pins. Then, a fully convolutional network, IREDGe, utilizing these additional features for IR drop map generation. 
In contrast, our paradigm manifests an obvious difference with this work. 
Firstly, PDNGraph represents the PDN structure using a directed graph. We take into account the specific locations of the PDN on the layout and bridge finely cell-PDN physical relation during the graph construction. In cases where the PDN of a chip is relatively uniform, the PDN density map would become a uniform map which powerless to capture the PDN. 
Secondly, the edges direction in PDNGraph simulating the current flow from the PDN structure to the cells. The graph branch in PDNNet can favorably perceive and directionally aggregate the features of neighboring features, which cannot be achieved by CNNs.

\section{Experiments}\label{sec:experiments}
In this section, we first conduct experiments on large public datasets to evaluate the performance of our model.
Then, we perform extra experimental ablation studies to explore the performance gained from several key designs.

\subsection{Experimental Setting}
\noindent\textbf{Environment Setups.} We implement our method using PyTorch and the DGL graph learning framework.
The experiments are conducted on a Linux machine with an NVIDIA A800 GPU, two 32-core Intel Xeon Processors (Ice Lake), and 1024GB memory.

\begin{table*}
\begin{center}
\caption{\small Evaluation results on CircuitNet dataset. 
Our PDNNet achieves consistent performance gain on five evaluation metrics in both large and small datasets. 
$\downarrow$ means “lower is better”, $\uparrow$ means “higher is better”.}
\label{table:main}
\vspace{0.3em}
\setlength{\tabcolsep}{1.0pt}
\scalebox{0.95}{
    \centering
    \setlength{\tabcolsep}{1mm}{
    \begin{tabular}{l | c c c c c c c c}
    \toprule[0.12em]
    \multirow{2}{*}{\textbf{Method}} & \multicolumn{7}{c}{\textbf{CircuitNet~\citep{chai2022circuitnet}}}  \\
         & NMAE~$\textcolor{black}{\downarrow}$ & $R^2$~$\textcolor{black}{\uparrow}$ & 
         PSNR~$\textcolor{black}{\uparrow}$ & SSIM~$\textcolor{black}{\uparrow}$ &
         Pear~$\textcolor{black}{\uparrow}$ & Spea~$\textcolor{black}{\uparrow}$ & Kend~$\textcolor{black}{\uparrow}$  & AUC~$\textcolor{black}{\uparrow}$ \\
         \cline{1-9}
        PowerNet~\citep{xie_powernet_2020} & 0.149  & 0.32 & 11.60 & 0.56 & 0.77 & 0.83 & 0.70 & 0.503 \\ 
        MAVIREC~\citep{chhabria_mavirec_2021} & 0.039 & 0.81 & 18.27 & 0.68& 0.91 & 0.85 & 0.77  & 0.942\\ 
        MAVIREC+image-based PDN feature & 0.037& 0.82 & 18.55 & 0.70 & 0.91 & 0.86 & 0.78 & 0.946\\ 
        \midrule[0.12em]
        \textbf{PDNNet} (Ours) &
         \textbf{0.028} & \textbf{0.84} & \textbf{19.35} & \textbf{0.72} & \textbf{0.92} & \textbf{0.87} & \textbf{0.81}  & 0.953 \\ 
         
        \bottomrule[0.12em]
    \end{tabular}}}
\end{center}
\end{table*}


\noindent\textbf{Datasets.} CircuitNet~\citep{chai2022circuitnet, chai2023circuitnet} is the large-scale public dataset in development that provides IC design used in the real industrial scenario.
Unlike most previous work using small datasets for validation, CircuitNet-N28 (28nm version) provides more than 10,000 samples from 6 basic RTL designs (54 different synthesized netlists) with a wide variety of macros (3 types), frequencies (50/200/500 MHz), and back-end flow (variant in settings of utilization, macro placement, power mesh, filler insertion, etc), which is favorably appropriate to examine the actual performance of dynamic IR drop prediction models.
We follow the same protocol as CircuitNet \citep{chai2022circuitnet}, 
to split the training and testing sets. That is, 4 designs (\texttt{RISCY-a, RISCY-b, RISCY-FPU-a, RISCY-FPU-b}) with 7078 samples are for training, and the rest 2 designs (\texttt{zero-riscy-a} and \texttt{zero-riscy-b}) with 3164 samples are for testing.
CircuitNet-N28 leverages Innovus and Voltus flow to obtain vectorless dynamic IR drop solutions.
In our experiment, we use the power feature that was initially provided in CirucitNet which incorporates 5 types of power patterns with 24 channels.
Further, we extract the PDN features from the Design Exchange Format (DEF) files while incorporating the power information to complete our PDNGraph construction.
We also use the whole dataset for conducting the ablation studies in~\Cref{sec:experiments:ablation}. 
Notice that the above training and validation datasets have no intersection across designs.
Refer to~\Cref{table:statistics} for more details.


\begin{figure}[!t]
\begin{center}
\setlength{\tabcolsep}{1.5pt}
\scalebox{0.9}{
\begin{tabular}[b]{c c c}

\includegraphics[width=.15\textwidth,valign=t]{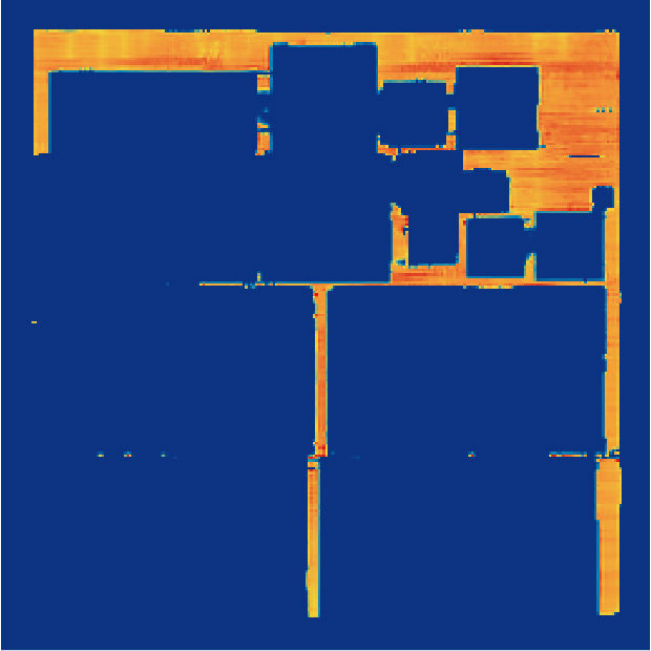} &   
\includegraphics[width=.15\textwidth,valign=t]{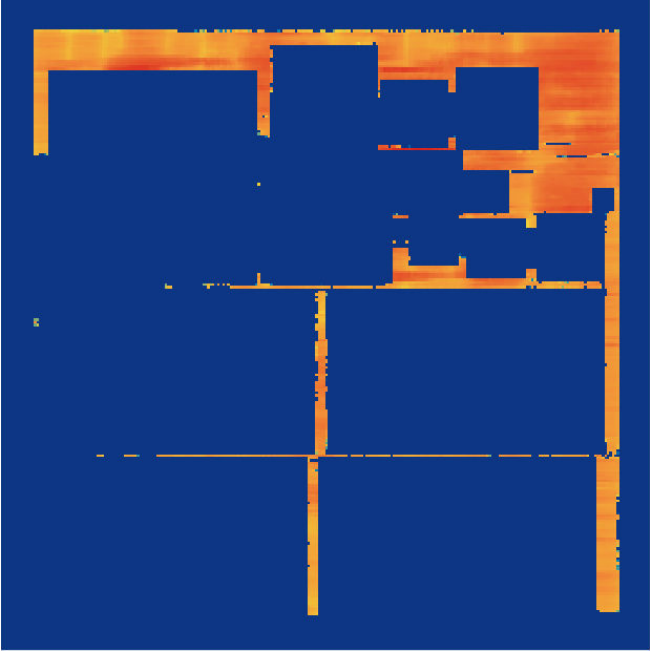} &   
\includegraphics[width=.15\textwidth,valign=t]{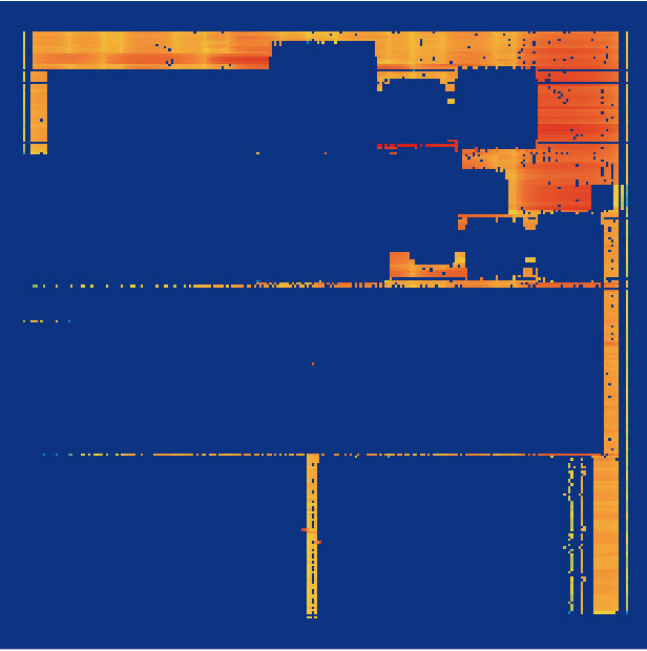}   
\\
\\
\includegraphics[width=.15\textwidth,valign=t]{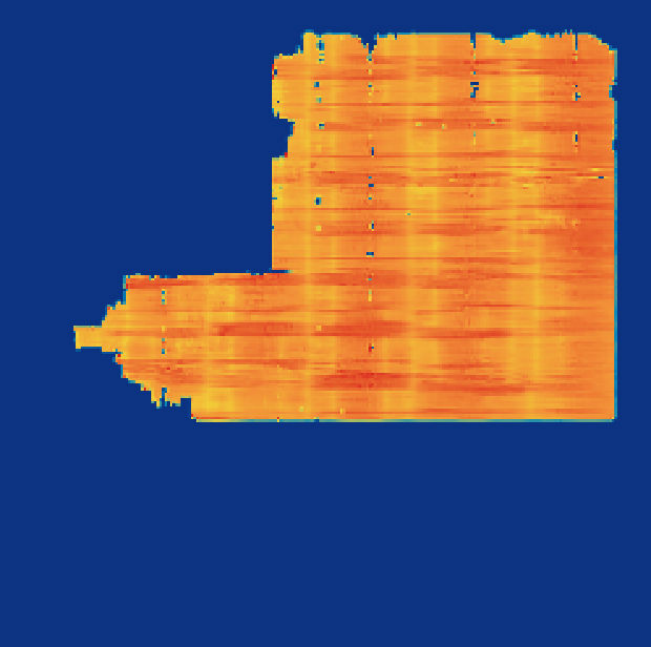} &   
\includegraphics[width=.15\textwidth,valign=t]{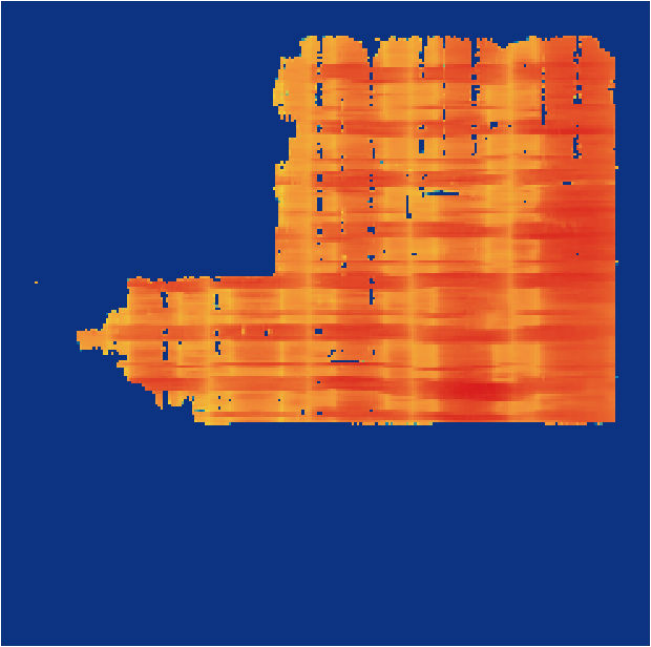} &   
\includegraphics[width=.15\textwidth,valign=t]{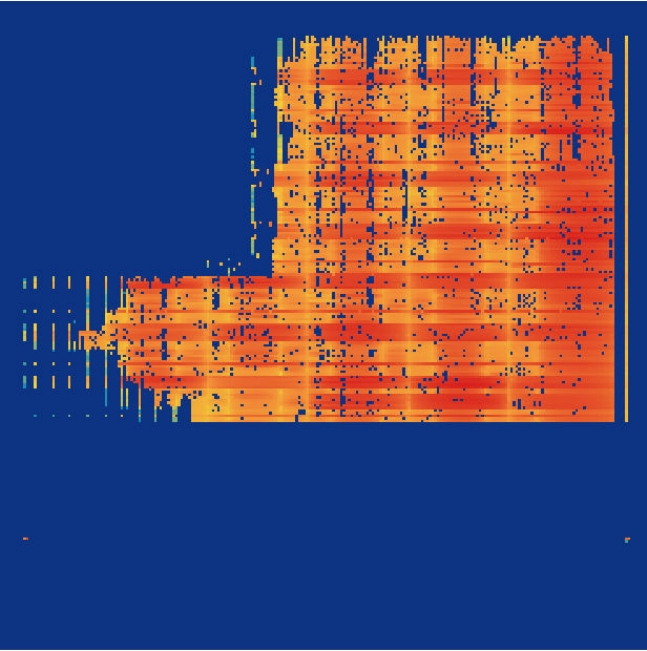}   
\\
\\
\ \  \small  \ MAVIREC~\citep{chhabria_mavirec_2021} & \small  Our PDNNet  & \small \   Ground Truth
\vspace{1mm} 
\end{tabular}
}
\end{center}
\caption{Qualitative evaluation of CircuitNet dataset. 
Our PDNNet generates IR drop prediction map with high structural fidelity, which demonstrates the effectiveness of our method. 
}
\label{fig:visual}
\end{figure}



\begin{table*}
\begin{center}
\caption{\small Results of constructing GNN branch with different message-passing layers.}
\label{table:gnn}
\vspace{0.3em}
\setlength{\tabcolsep}{2.0pt}
\scalebox{1.1}{
    \centering
    \begin{tabular}{l c c c c c c c}
    \toprule[0.12em]
        Network & NMAE~$\textcolor{black}{\downarrow}$ & $R^2$~$\textcolor{black}{\uparrow}$ & 
         PSNR~$\textcolor{black}{\uparrow}$ & SSIM~$\textcolor{black}{\uparrow}$ &
         Pear~$\textcolor{black}{\uparrow}$ & Spea~$\textcolor{black}{\uparrow}$ & Kend~$\textcolor{black}{\uparrow}$ \\ 
        \midrule[0.12em]
        Full Voltage-Drop Acquiring Layers &  0.0286 & 0.8343 & 19.2704 & 0.7156 & 0.9173 & 0.8562 & 0.7902
        \\ 
        Full Neighbor Influence Layers & 0.0287 & 0.8316 & 19.2119 & 0.7151 & 0.9166 & 0.8492 & 0.7832 \\ 
        \midrule
        \textbf{Mixed GNN Layers (our GNN branch)} & \textbf{0.0282} & \textbf{0.8369} & \textbf{19.3483} & \textbf{0.7190} & \textbf{0.9186} & \textbf{0.8709} & \textbf{0.8064} \\ 
        \bottomrule[0.12em]
    \end{tabular}}
\end{center}
\end{table*}
 

\begin{table*}
\begin{center}
\caption{\small Influence of GNN \& CNN branch of PDNNet. 
}
\label{table:dual-branch}
\vspace{0.3em}
\setlength{\tabcolsep}{2.0pt}
\scalebox{1.1}{
    \centering
    \begin{tabular}{l| c c c|c c c c c c c}
    \toprule[0.12em]
        \textbf{Network} & Single-Branch & Dual-Branch &Heterogeneous&    
        NMAE~$\textcolor{black}{\downarrow}$ & $R^2$~$\textcolor{black}{\uparrow}$ & 
         PSNR~$\textcolor{black}{\uparrow}$ & SSIM~$\textcolor{black}{\uparrow}$ &
         Pear~$\textcolor{black}{\uparrow}$ & Spea~$\textcolor{black}{\uparrow}$ & Kend~$\textcolor{black}{\uparrow}$ \\ 
        \midrule[0.12em]
        \multicolumn{1}{l|}{\multirow{2}{*}{CNN}} &\color{violet}{\CheckmarkBold}         & \color{gray}{\XSolidBrush}   &  \color{gray}{\XSolidBrush}    & 0.0370 & 0.8100 & 18.3112  & 0.6956 & 0.9130 &  0.8646 &  0.7890  \\ 
          &   \color{gray}{\XSolidBrush}       & \color{violet}{\CheckmarkBold}       & \color{gray}{\XSolidBrush}  & 0.0358 & 0.8193 & 18.8091 &  0.7004 &0.9122 & 0.8641 & 0.7904 \\ 
        \midrule
        \multicolumn{1}{l|}{\multirow{2}{*}{GNN}}                       & \color{violet}{\CheckmarkBold}        & \color{gray}{\XSolidBrush}    &  \color{gray}{\XSolidBrush}  & 0.0365 & 0.8120 & 18.7100 & 0.6897 &0.9066 & 0.8865 & 0.7911 \\ 
        &      \color{gray}{\XSolidBrush}    &  \color{violet}{\CheckmarkBold}     &  \color{gray}{\XSolidBrush} & 0.0336 & 0.8231 & 18.9977  & 0.6983 & 0.9119 & 0.8690 & 0.7816  \\ 
        \midrule
        \textbf{PDNNet}  &    \color{gray}{\XSolidBrush}  & \color{violet}{\CheckmarkBold} & \color{violet}{\CheckmarkBold} & \textbf{0.0282} & \textbf{0.8369} & \textbf{19.3483} & \textbf{0.7190} & \textbf{0.9186} & \textbf{0.8709} & \textbf{0.8064} \\ 
        \bottomrule[0.12em]
    \end{tabular}}
\end{center}
\end{table*}
 

\noindent\textbf{Implementation Details.}
We train our model, using Adam optimizer with $\beta_1{=}0.9$, $\beta_2{=}0.999$, the initial learning rate is set as 0.0008 and apply a low weight decay of $0.0001$. 
Other baseline methods are all trained to achieve convergence.
We gradually reduced the initial learning rate with the cosine annealing strategy. 
Datasets pre-processing follows~\citep{chai2022circuitnet}. 
Note that we do not impose any data augmentation transform during the training stage.
As for the training loss, we adopt L1 and Dice (Dice similarity coefficient) ~\citep{dice} loss and treat them with equal weight.
Note that our modified network outputs an IR drop map consistent in dimensions with the layout, enhancing its validity. For a fair comparison, all baseline methods are modified to adopt different size output.


\subsection{Main Experiment}
We first compare our PDNNet with the state-of-the-art CNNs from the literature.
The first baseline for comparison is PowerNet~\citep{xie_powernet_2020}, which is the first paper to perform dynamic IR drop prediction with deep-learning method.
The second is MAVIREC~\citep{chhabria_mavirec_2021}, a UNet-like model learning on dynamic IR drop with input vector, which is currently the state-of-the-art among the vector-based or vector-less methods.
The last is incorporate PDN density map and power pad effective distance map from  IREDGe\citep{static-irdrop} as additional features into the MAVIREC network. This method serves as a comparison with our GNN approach for perceiving the PDN structure.
In general, the CircuitNet dataset expects neural networks to have strong generalization and representation capabilities due to its diverse design compositions.
Thus, for evaluation metrics, we report both the accuracy metric (\emph{i.e.,} Normalized Mean Absolute Error (NMAE)), coefficient of determination (\emph{i.e.}, R-square ($R^{2}$)), image similarity (\emph{i.e.}, Peak Signal-to-Noise Ratio, Structural Similarity Index Measure), and the graph evaluation criterion (\emph{i.e.}, Pearson, Spearman, Kendall correlation coefficient), and area under the curve (AUC) to comprehensively assess the models' performance.
We abbreviate Pearson to \emph{Pear} for shorten, Spearman and Kendall convert to \emph{Spea} and \emph{Kend} as the same.
\emph{PSNR} and \emph{SSIM} are commonly used terms to refer to the above image similarity indexes.
%
We calculate NMAE metric as $\frac{1}{N}\sum_{i=0}^{{N}-1}\frac{{\left| y_{i}-\tilde{y}_{i} \right|}}{max(y_{i})-min(y_{i})}$.
%
%

\noindent\textbf{Accuracy Comparison. } 

The~\Cref{table:main} shows the results.
Compared to the PowerNet, PDNNet achieves improvement (\textbf{0.028} \emph{vs.} 0.149) in NMAE metrics, which depicts the incomparable generalization of our PDNNet, other metrics also support this viewpoint.
The weak performance of PowerNet is primarily due to its simplistic network design and non-mainstream 3D learning strategy, which limits its ability to learn deeper features related to IR drop.
The typical CNN architecture, MAVIREC, still has a great margin compared with our PDNNet (0.039 \emph{vs.} \textbf{0.028} in NMAE, 0.81 \emph{vs.} \textbf{0.84} in $R^{2}$, 18.27dB \emph{vs.} \textbf{19.35dB} in PSNR, 0.68 \emph{vs.} \textbf{0.72} in SSIM, 0.91 \emph{vs.} \textbf{0.92} in Pearson, etc).
When comparing with image-based PDN representation methods. PDNNet also present superiority in all metrics. 
It demonstrates the feature extraction process of GNNs for PDN structures can be viewed as a learnable representation, offering superior expressiveness compared to fixed image-based features.
It is worth noting that our PDNNet has consistent and significant performance gains over all seven metrics, which demonstrates the complementary role of PDN attributes perception and dynamic current and during IR drop representation.
%
%
Visualization comparison results are shown in~\Cref{fig:visual}. 
The figures from the left to right columns are results from MAVIREC~\citep{chhabria_mavirec_2021}, results from our PDNNet, and ground truth label.
The comparison results clearly show that our PDNNet generates IR drop prediction map with high structural fidelity, which demonstrates the effectiveness of our method.
%

\noindent\textbf{AUC Comparison. } 

We also calculate the AUC score to measure model representation performance, PDNNet achieves \textbf{0.953}, outperforming the result of 0.946 in MAVIREC+image-based feature, 0.942 in MAVIREC, and 0.503 in PowerNet.


\subsection{Ablation Studies}\label{sec:experiments:ablation}
In this section, we investigate the effectiveness of each individual technique by ablation studies. 
We perform two series of experiments to investigate the following research questions:
\emph{\textbf{RQ1}. How effective is the mixure of different message-passing layers in the GNN branch compared with each individual of them?} 
\emph{\textbf{RQ2}. How effective is the dual-branch heterogeneous PDNNet compared with the pure CNN or GNN methods?}
%



\noindent\textbf{Physical Intuitive GNN Branch Design (RQ1). }

We perform a preliminary exploration of GNN branch design by constructing two variants, a full layer of voltage-drop acquiring blocks and a full layer of neighbor influence blocks.
We train the above models and PDNNet on the whole CircuitNet with the same settings.
Results in~\Cref{table:gnn}, show that two variants have similar performance. 
The branch with full neighbor influence layers is slightly higher than its voltage-drop acquiring counterpart in NMAE metric (0.0287 \emph{vs.} 0.0286), while lower in $R^{2}$ (0.8316 \emph{vs.} 0.8343).
However, by merging these distinctive layers in a specific manner, we observe a performance improvement in our PDNNet's GNN branch over all metrics.
These results reinforce the rationality of the formulation in~\Cref{sec:preliminary:idrop} that shallow layers of GNN branch utilize current-related features to learn the preliminary representations of voltage drop, while deeper layers are to refine nodes' voltage drop features influenced by their neighbors.

\noindent\textbf{Dual-Branch Heterogeneous Architecture Analysis (RQ2).} 
We perform a controlled study to assess PDNNet's performance versus each variant model's. 
The model set includes single-branch CNN and dual-branch CNN, plus single-branch GNN and dual-branch GNN. 
The dual-branch variants stand for using the same architecture as PDNNet while simply substituting with two identical branches (CNN or GNN).
\Cref{table:dual-branch} contains the comparison of the results.
We have the following observations. 
First, although CNNs dominate the IR drop prediction in previous works, we interestingly observe that our GNN branch's performance in a single branch has similarly performance compared its CNN-based counterpart, whilst with a simple PDN perception and relatively shallow layers (two GNN layers, each accompanied by one GNN block. The parameter is also less than CNN, 41.15K vs. 17.07M).
This phenomenon is largely due to our specifically designed PDNGraph data structure which effectively mimics current flow and fine-grained cell-PDN relation within the PDN grid.
In dual-branch architecture, despite the NMAE, $R^2$ and PSNR, the dual-branch CNN have similar performance compared with GNN counterpart in other metrics.However, the parameter in dual-branch CNN is far more than GNN 34.13M vs. 78.02K.
Second, compared to a single-branch, dual-branch architecture indeed boosts the prediction performance.
However, both of these variants do not outperform our heterogeneous PDNNet (0.0358/0.0336 \emph{vs.} \textbf{0.0282}, and 0.8193/0.8231 \emph{vs.} \textbf{0.8396}) with 17.1M parameters, which demonstrate the PDN configuration and dynamic current pattern representations are two complementary aspects for effective IR drop prediction.

\section{Conclusion}\label{sec:conclusion}
This paper introduces a novel approach to accurately predict dynamic IR drop on the power delivery network in IC design.
We present a novel graph structure, PDNGraph, to properly represent the PDN configuration and the fine-grained cell-PDN relation.
Moreover, we propose a novel network, PDNNet, a heterogeneous network with GNN-CNN branches to favorably perceive the above crucial features.
Extensive experiments on a large public dataset demonstrate that our PDNNet outperforms the existing state-of-the-art CNN-based methods by a large margin and achieves a 545$\times$ speedup compared to the commercial tool with minimal error.
We hope our work can foster further inspiration for EDA researchers.


{
\small
\bibliographystyle{IEEEtran}
\bibliography{./Ref/Top,./Ref/PD,./Ref/Software,./Ref/ALG,./Ref/bib,./Ref/bib2}
}

\end{document}